\documentclass[a4paper,conference]{IEEEtran}
\ifCLASSINFOpdf
\else
\fi
\usepackage{times}
\usepackage{soul}
\usepackage{url}
\usepackage[hidelinks]{hyperref}
\usepackage[utf8]{inputenc}
\usepackage[small]{caption}
\usepackage{graphicx}
\usepackage{amsmath}
\usepackage{booktabs}
\usepackage{algorithm}
\usepackage{algorithmic}
\usepackage{tikz}
\usepackage{ifthen}
\usepackage{xcolor}
\definecolor{darkblue}{HTML}{1F4E79}
\definecolor{lightblue}{HTML}{00B0F0}
\definecolor{salmon}{HTML}{FF9C6B}
\usepackage{subfig}
\usetikzlibrary{spy}
\usetikzlibrary{shapes.geometric, arrows, backgrounds, fit}
\tikzstyle{layer} = [rectangle, minimum width=1cm, minimum height=0.5cm,text centered, draw=black, fill=lightblue]
\tikzstyle{cell} = [rectangle, minimum width=1cm, minimum height=0.5cm,text centered, draw=black, fill=salmon]
\tikzstyle{darklayer} = [rectangle, minimum width=0.5cm, minimum height=0.5cm,text centered, draw=black, fill=darkblue]
\tikzstyle{state} = [circle, inner sep=0pt, minimum width=0.5cm, minimum height=0.5cm,text centered, draw=black, fill=darkblue]
\tikzstyle{arrow} = [thick,->,>=stealth]

\begin{document}
%
\title{Fast, Accurate and Lightweight Super-Resolution with Neural Architecture Search}

\author{\IEEEauthorblockN{Xiangxiang Chu, Bo Zhang, Hailong Ma, Ruijun Xu}
\IEEEauthorblockA{Xiaomi AI Lab\\
Beijing, China \\
Email: \{chuxiangxiang,zhangbo11,mahailong,xuruijun\}@xiaomi.com}
\and
\IEEEauthorblockN{Qingyuan Li}
\IEEEauthorblockA{Xiaomi IoT\\
Beijing, China\\
Email: liqingyuan@xiaomi.com}
}


%


\maketitle

\begin{abstract}
Deep convolutional neural networks demonstrate impressive results in the super-resolution domain. A series of studies concentrate on improving peak signal noise ratio (PSNR) by using much deeper layers, which are not friendly to constrained resources. Pursuing a trade-off between the restoration capacity and the simplicity of models is still non-trivial. Recent contributions are struggling to manually maximize this balance, while our work achieves the same goal automatically with neural architecture search. Specifically, we handle super-resolution with a multi-objective approach. We also propose an elastic search tactic at both micro and macro level, based on a hybrid controller that profits from evolutionary computation and reinforcement learning.
Quantitative experiments help us to draw a conclusion that our generated models dominate most of the state-of-the-art methods with respect to the individual FLOPS.  
\end{abstract}


%
\IEEEpeerreviewmaketitle

\section{Introduction and Related Work}

As a classical task in computer vision, single image super-resolution (SISR) is aimed to restore a high-resolution image from a degraded low-resolution one, which is known as an ill-posed inverse procedure.  Most of the recent works on SISR have shifted their approaches to deep learning, and they have surpassed other SISR algorithms with big margins \cite{dong2014learning,kim2016accurate,he2016deep,ahn2018fast}.


Nonetheless, these human-designed models are tenuous to fine-tune or to compress. Meantime, neural architecture search has produced dominating models in classification tasks \cite{zoph2016neural,zoph2017learning}. Following this trend, a novel work by \cite{chu2019multi} has shed light on the SISR task with a reinforced evolutionary search method, which has achieved results outperforming some notable networks including VDSR \cite{kim2016accurate}. We are distinct to \cite{chu2019multi} by stepping forward to design a dense search space which allows searching in both macro and micro level, which has led to significantly better visual results.

In this paper, we dive deeper into the SISR task with elastic neural architecture search, hitting a record comparable to CARN and CARN-M \cite{ahn2018fast} \footnote{Our models are released at \url{https://github.com/falsr/FALSR}.}. Our main contributions can be summarized in the following four aspects,
\begin{itemize}
	\item releasing several fast, accurate and lightweight super-resolution architectures and models (FALSR-A being the best regarding visual effects), which are highly competitive with recent state-of-the-art methods,
	\item performing elastic search by combining micro and macro space on the cell-level to boost capacity, 
	\item building super-resolution as a constrained multi-objective optimization problem and applying a hybrid model generation method to balance exploration and exploitation,
	\item producing high-quality models that can meet various requirements under given constraints within a single run.
\end{itemize}

\begin{figure*}
	\vskip 0.35in
	\noindent\resizebox{\textwidth}{!}{
	\begin{tikzpicture}
		\draw[use as bounding box, transparent] (-1.8,-1.8) rectangle (17.2, 3.2);


		\newcommand{\networkLayer}[7]{
			\def\a{#1} 
			\def\b{0.02}
			\def\c{#2} 
			\def\t{#3} 
			\def\d{#4} 
			\filldraw[#5] (\t+\b,\b,\a+\d) -- (\c+\t-\b,\b,\a+\d) -- (\c+\t-\b,\a-\b,\a+\d) -- (\t+\b,\a-\b,\a+\d) -- (\t+\b,\b,\a+\d); 
			\filldraw[#5] (\t+\b,\a,\a-\b+\d) -- (\c+\t-\b,\a,\a-\b+\d) -- (\c+\t-\b,\a,\b+\d) -- (\t+\b,\a,\b+\d);

			\ifthenelse {\equal{#5} {}}
			{} 
			{\filldraw[#5] (\c+\t,\b,\a-\b+\d) -- (\c+\t,\b,\b+\d) -- (\c+\t,\a-\b,\b+\d) -- (\c+\t,\a-\b,\a-\b+\d);} 
		
			\draw[line width=0.25mm,#7](\c+\t,0,\d) -- (\c+\t,\a,\d) -- (\t,\a,\d);                                                      
			\draw[line width=0.25mm,#7](\t,0,\a+\d) -- (\c+\t,0,\a+\d) node[midway,below] {#6} -- (\c+\t,\a,\a+\d) -- (\t,\a,\a+\d) -- (\t,0,\a+\d); 
			\draw[line width=0.25mm,#7](\c+\t,0,\d) -- (\c+\t,0,\a+\d);
			\draw[line width=0.25mm,#7](\c+\t,\a,\d) -- (\c+\t,\a,\a+\d);
			\draw[line width=0.25mm,#7](\t,\a,\d) -- (\t,\a,\a+\d);
			}

		\begin{scope}[yslant=1,cm={0.5,0,0,1,(0,0.5)}]
			\node [transform shape](input image) at (0,0) {\includegraphics[scale=.2]{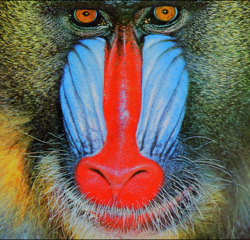}};
		\end{scope}
		\networkLayer{2.0}{0.3}{2}{0}{color=lightblue}{Feature Extractor}{}    
		\networkLayer{2}{0.5}{3}{-2.5}{color=salmon}{cell $1$}{}        
		\networkLayer{2}{0.5}{5}{-2.5}{color=salmon}{cell $2$}{}
		\networkLayer{2}{0.5}{7}{-2.5}{color=salmon!50}{}{dashed}
		\networkLayer{2}{0.5}{9}{-2.5}{color=salmon}{cell $n$}{}
		\networkLayer{2.0}{0.3}{12}{0.0}{color=lightblue}{}{}   
		\networkLayer{3.0}{0.02}{13.6}{0.0}{color=lightblue}{Subpixel Upsampling}{}  
		\begin{scope}[yslant=1,cm={0.5,0,0,1,(9.2,-13.2)}]
			\node [transform shape](input image) at (11,0) {\includegraphics[scale=.2]{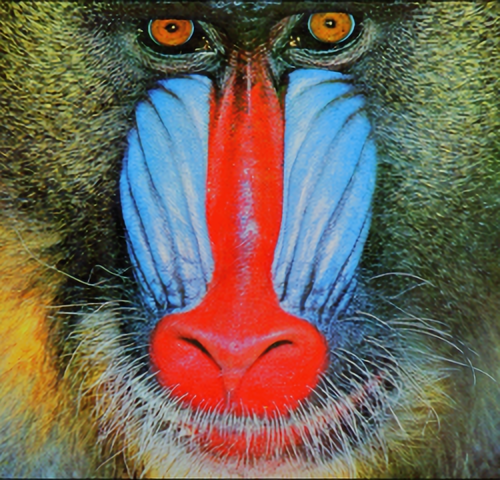}};
		\end{scope}

		\node  (feature) at (2,2) {}; 
		\node  (extractor) at (2,-0.2) {}; 
		\node  (cell_0) at (4,3) {};
		\node  (cell_1) at (6,3) {}; 
		\node  (cell_2) at (8,3) {}; 
		\node  (cell_3) at (10,3) {}; 
		\node  (subpixel) at (12,2) {}; 
		\node  (upsample) at (11,0) {}; 
		\draw [draw=darkblue,arrow] (feature) edge [bend left=35] (cell_1);
		\draw [draw=darkblue,arrow] (feature) edge [bend left=45] (cell_2);
		\draw [draw=darkblue,arrow] (extractor) edge [bend left=-15] (upsample);
		\draw [draw=salmon, arrow] (cell_0) edge [bend left=25] (cell_2);
		\draw [draw=salmon, arrow] (cell_0) edge [bend left=25] (cell_3);
		\draw [draw=salmon!80, arrow] (cell_2) edge [bend left=45] (subpixel);
	\end{tikzpicture}
	}
	\vskip 0.1in
	\caption{Neural Architecture of Super-Resolution (the arrows denote skip connections). }
	\label{fig:nas-sr}
\end{figure*}
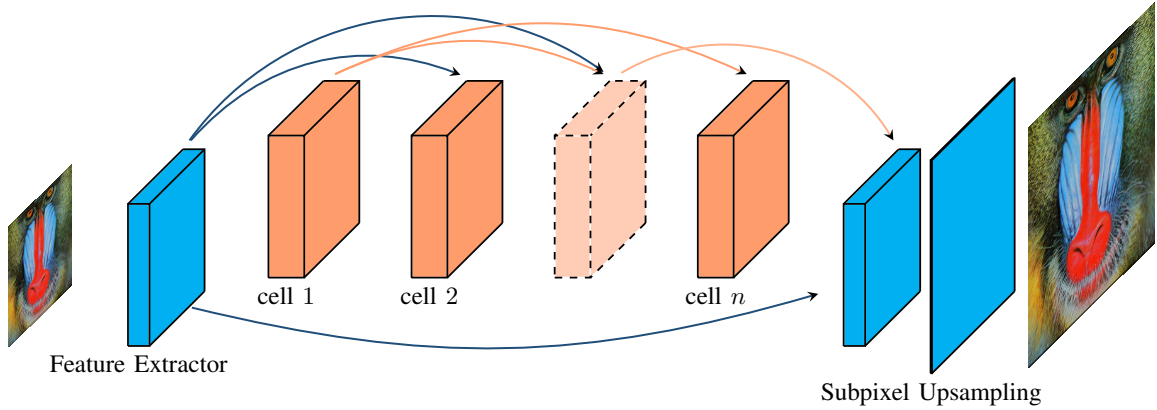

\section{Pipeline Architecture}
Like most of Neural Architecture Search (NAS) approaches, our pipeline contains three principle ingredients: an elastic search space, a hybrid model generator and a model evaluator based on incomplete training. It is explained in detail in the following sections.

Similar to \cite{lu2018nsga,chu2019multi}, we also apply NSGA-II \cite{deb2002fast} to solve the multi-objective problem. Our work differs from them by using a hybrid controller and a cell-based elastic search space that enables both macro and micro search. 

We take three objectives into account for the super-resolution task,
\begin{itemize}
	\item quantitative metric to reflect the performance of models (PSNR),
	\item quantitative metric to evaluate the computational cost of each model (mult-adds),
	\item number of parameters.
\end{itemize}
In addition, we consider the following constraints,
\begin{itemize}
	\item minimal PSNR for practical visual perception,
	\item maximal mult-adds regarding resource limits.
\end{itemize}

\section{Elastic Search Space}

Our search space is designed to perform both micro and macro search. The former is used to choose promising cells within each cell block, which can be viewed as a feature extraction selector. In contrast, the latter is aimed to search backbone connections for different cell blocks, which plays a role of combining features at selected levels. In addition, we use one cell block as our minimum search element for two reasons: design flexibility, and broad representational capacity. 

Typically, the super-resolution task can be divided into three sub-procedures: feature extraction, nonlinear mapping, and restoration. Since most of the deep learning approaches concentrate on the second part, we design our search space to describe the mapping while fixing others. Figure \ref{fig:nas-sr} depicts our main flow for super-resolution. Thus, a complete model contains a predefined feature extractor (a 2D convolution with 32 3 $\times$ 3 filters), $n$ cell blocks drawn from the micro search space which are joined by the connections from macro search space, and subpixel-based upsampling and restoration\footnote{Our upsampling contains a 2D convolution with 32 3$\times $3 filters, followed by a 3 $\times$ 3 convolution with one filter of unit stride.}.

\subsection{Cell-Level Micro Search Space}
For simplicity, all cell blocks share the same cell search space $S$. In specific, the micro search space comprises the following elements:
 
 \begin{itemize}
 	\item convolutions: 2D convolution, grouped convolution with groups in $\{2, 4\}$, inverted bottleneck block with an expansion rate of $2$,
 	\item channels: $\{16, 32, 48, 64\}$,
 	\item kernels: \{1, 3\},
 	\item in-cell residual connections:$\{ \text{True}, \text{False}\}$,
 	\item repeated blocks:$\{1, 2, 4\}$.
 \end{itemize}
 Therefore, the size of  micro space for $n$ cell blocks is $192^n$.
 
\subsection{Intercell Macro Search Space}
The macro search space defines the connections among different cell blocks. Specifically, for the $i$-th cell block $CB_i$, there are $n + 1 - i$  choices of connections to build the information flow from the input of $CB_i$ to its following cell blocks\footnote{Here, $i$ starts with 1.}. Furthermore, we use $c_i^j$ to represent the path from input of $CB_i$ to $CB_j$. We set  $c^j_i =1 $ if there is a connection path between them, otherwise 0. Therefore, the size of  macro space for $n$ cell blocks is $2^{n(n+1)/2}$. In summary, the size of the total space is $192^n \times 2^{n(n+1)/2}$.
 
\section{Model Generator}
Our model generator is a hybrid controller involving both reinforcement learning  (RL) and an evolutionary algorithm (EA). The EA part handles the iteration process and RL is used to bring exploitation.
To be specific, the iteration is controlled by NSGA-II \cite{deb2002fast}, which contains four sub-procedures:  population initialization, selection, crossover, and mutation. To avoid verbosity, we only cover our variations to NSGA-II.

\subsection{Model Meta Encoding}
One model is denoted by two parts: forward-connected cells and their information connections. We use the indices of operators from the operator set to encode the cells, and a nested list to depict the connections. Namely, given a model $M$ with $n$ cells, its corresponding chromosome can be depicted by ($M_{mic}, M_{mac}$), where $M_{mic}$ and $M_{mac}$ are defined as follows,
\begin{align}
\begin{split}
M_{mic} &= (x_1, x_2, ..., x_n)
\end{split}\\
\begin{split}
M_{mac} &= (c_1^{1:n}, c_2^{2:n}, ..., c_n^{n}) \\
c_i^{i:n} &= (c_i^i, c_i^{i+1}, .. c_i^{n})
\end{split}
\end{align}

\subsection{Initialization}

We begin with $N$ populations and we emphasize the diversities of cells. In effect, to generate a model,  we randomly sample a cell from $S$ and repeat it for $n$ times. In case $N$ is larger than the size of $S$, models are arbitrarily sampled without repeating cells.

As for connections, we sample from a categorical distribution. While in each category, we pick uniformly,  i.e. $p \sim \mathcal{U} (0, 1)$. To formalize, the connections are built based on the following rules,
\begin{equation}
\left\{
\begin{array}{lr}
\text{random connections} &  0 \leq p < p_r \\
\text{dense connections} &  p_r \leq p < p_r+p_{den} \\
\text{no connnections} & p_r+p_{den} \leq p < 1
\end{array}
\right.
\end{equation}

\subsection{Tournament Selection}
We calculate the crowding distance as noted in \cite{chu2018improved} to render a uniform distribution of our models, and we apply tournament selection ($k=2$) to control the evolution pressure.
\subsection{Crossover}
To encourage exploration, single-point crossovers are performed simultaneously in both micro and macro space.  Given two models $A$ ($M_{mic(A)}, M_{mac(A)}$) and $B$ ($M_{mic(B)}, M_{mac(B)}$), a new chromosome $C$  can be generated as,
\begin{equation}
\begin{split}
	M_{mic(C)} &=  (x_{1A}, x_{2A}, ...,x_{iB},..., x_{nA})\\
	M_{mac(C)} &= (c^{1:n}_{1A}, c^{2:n}_{2A}, ..., c^{j:n}_{jB},...,c^{n}_{nA})\\ 
\end{split}
\end{equation}
where $i$ and $j$ are chosen positions respectively for micro and macro genes. Informally, the crossover procedure contributes more to exploitation than to exploration.

\subsection{Mutation}
We again apply a categorical distribution to balance exploration and exploitation.
\subsubsection{Exploration}
To encourage exploration, we combine random mutation with roulette wheel selection (RWS). Since we treat super-resolution as a multi-objective problem, FLOPS and the number of parameters are two objectives that can be evaluated soon after meta encodings are available.  In particular, we also sample from a categorical distribution to determine mutation strategies, i.e. random mutation (with an upper-bound probability $p_{mr}$) or mutated by roulette wheel selection to handle FLOPS (lower than $p_{mf}$) or parameters. Formally,
\begin{equation}
\left\{
\begin{array}{lr}
\text{random mutation} &  0 \leq p < p_{mr} \\
\text{RWS for FLOPS}  &  p_{mr} \leq p < p_{mf} \\
\text{RWS for params} &  p_{mf}  \leq p < 1
\end{array}
\right.
\end{equation}
Whenever we need to mutate a model $M$ by RWS, we keep $M_{mac}$ unchanged. Since each cell shares the same operator set $S$, we perform RWS on $S$ for $n$ times to generate $M_{mic}$. Strictly speaking, given $M_{mac}$, it's intractable to execute a complete RWS (involving $192^n$ models). Instead, it can be approximated based on $S$ (involving $192$ basic operators). Besides, we scale FLOPS and the number of parameters logarithmically before RWS. 

%

\subsubsection{Exploitation}
To enhance exploitation, we apply a reinforcement driven mutation.

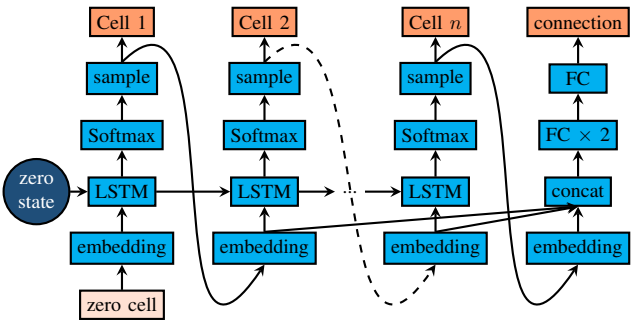
\begin{figure}[ht]
	\vskip -0.2in
	\begin{center}
		\centerline{
			\vspace*{-45pt} 
			\begin{tikzpicture}[thick,scale=0.75, every node/.style={scale=0.75},node distance=2cm]
			\node (cell_1) [cell] {Cell 1};
			\node (sample_1) [layer, below of=cell_1,yshift=1cm] {sample};
			\node (softmax_1) [layer, below of=sample_1,yshift=1cm] {Softmax};
			\node (lstm_1) [layer, below of=softmax_1,yshift=1cm] {LSTM};
			\node (embed_1) [layer, below of=lstm_1,yshift=1cm] {embedding};
			\node (zero_1) [cell, below of=embed_1,yshift=1cm,fill=salmon!30] {zero cell};
			\node (start) [state, left of=lstm_1, xshift=0.5cm,text width=1cm] {\textcolor{white}{zero state}};
			\node (cell_2) [cell, right of=cell_1,xshift=0.5cm] {Cell 2};
			\node (sample_2) [layer, below of=cell_2,yshift=1cm] {sample};
			\node (softmax_2) [layer, below of=sample_2,yshift=1cm] {Softmax};
			\node (lstm_2) [layer, below of=softmax_2,yshift=1cm] {LSTM};
			\node (embed_2) [layer, below of=lstm_2,yshift=1cm] {embedding};
			\node (ellipsis) [right of=lstm_2,xshift=-0.5cm] {...};
			\node (cell_n) [cell, right of=cell_2,xshift=1cm] {Cell $n$};
			\node (sample_3) [layer, below of=cell_n,yshift=1cm] {sample};
			\node (softmax_3) [layer, below of=sample_3,yshift=1cm] {Softmax};
			\node (lstm_3) [layer, below of=softmax_3,yshift=1cm] {LSTM};
			\node (embed_3) [layer, below of=lstm_3,yshift=1cm] {embedding};
			\node (embed_4) [layer, right of=embed_3, xshift=0.5cm] {embedding};
			\node (concat) [layer, above of=embed_4, yshift=-1cm] {concat};
			\node (fc_2) [layer, above of=concat,yshift=-1cm] {FC $\times$ 2};
			\node (fc_3) [layer, above of=fc_2,yshift=-1cm] {FC};
			\node (conn) [cell, above of=fc_3,yshift=-1cm] {connection};
			\draw [arrow] (zero_1) -> (embed_1);
			\draw [arrow] (embed_1) -> (lstm_1);
			\draw [arrow] (lstm_1) -> (softmax_1);
			\draw [arrow] (softmax_1) -> (sample_1);
			\draw [arrow] (sample_1) -> (cell_1);
			\draw [arrow] (sample_1.north)[out=90] .. controls +(2,2) and +(-2,-3) .. (embed_2.south);
			\draw [arrow] (embed_2) -> (lstm_2);
			\draw [arrow] (lstm_2) -> (softmax_2);
			\draw [arrow] (softmax_2) -> (sample_2);
			\draw [arrow] (sample_2) -> (cell_2);
			\draw [arrow,dashed] (sample_2.north) [out=90] .. controls +(2,2) and +(-2,-3) .. (embed_3.south);
			\draw [arrow] (embed_3) -> (lstm_3);
			\draw [arrow] (lstm_3) -> (softmax_3);
			\draw [arrow] (softmax_3) -> (sample_3);
			\draw [arrow] (sample_3) -- (cell_n);
			\draw [arrow] (sample_3.north) [out=90] .. controls +(2,2) and +(-2,-3) .. (embed_4.south);
			\draw [arrow] (embed_4) -> (concat);
			\draw [arrow] (concat) -> (fc_2);
			\draw [arrow] (fc_2) -> (fc_3);
			\draw [arrow] (fc_3) -> (conn);
			\draw [arrow] (start) -> (lstm_1);
			\draw [arrow] (lstm_1) -> (lstm_2);
			\draw [arrow] (lstm_2)-> (ellipsis);
			\draw [arrow] (ellipsis)-> (lstm_3);
			\draw [arrow] (embed_2.north) to (concat.south);
			\draw [arrow] (embed_3.north) to (concat.south);
			\end{tikzpicture}
		}
		\vskip 0.1in
		\caption{The controller network to generate cells and connections.}
		\label{fig:rl_controlled_mutation}
	\end{center}
	\vskip -0.3in
\end{figure}

We use a neural controller to mutate, which is shown in Figure~\ref{fig:rl_controlled_mutation}. Specifically, the embedding features for $M_{mic}$ are concatenated, and then are injected into  3 fully-connected layers to generate $M_{mac}$. The last layer has $n(n+1)/2$ neutrons to represent connections, with its output denoted as $O^{mac}$.

The network parameters can be partitioned into two groups, $\theta^{mic}$ and $\theta^{mac}$. The probability of selecting $S_i$ for cell $j$ is $p(cell_i = S_i|\theta^{mic})$ and for the connection $c_i^j = 1$, we have $p(c_i^j=1|\theta^{mac})=O^{mac}_{(i-1)*(n+1-0.5*i)+j}$. Thus, the gradient $g(\theta)$ can be calculated as follows:
\begin{equation}
\begin{split}
	g(\theta) &= -\nabla_\theta [ \sum_{i=1}^{n}\log p(cell_i = S_i|\theta^{mic})*R_i+ \\
	&\sum_{j=1}^{n(n+1)/2} c_j\log O^{mac}_j*R_j+ \\
	&(1-c_j) \log(1- O^{mac}_j)*R_j].
\end{split}
\label{eq:pg}
\end{equation}
In Equation~\ref{eq:pg}, $R_i$ and $R_j$ are the discounted accumulated rewards. Here, we set the discount parameter $\gamma = 1.0$. 

\section{Evaluator}
The evaluator calculates the scores of the models generated by the controller. In the beginning,  we attempted to train an RNN regressor to predict the performances of models, with data collected in previous pipeline execution. However, its validation error is too high to continue. Instead, each model is trained for a relatively short time (see the `incomplete training' part in Section \ref{ref:setup}) to roughly differentiate various models. At the end of the incomplete training, we evaluate mean square errors on test datasets. 

\section{Experiments}
\subsection{Setup}\label{ref:setup}
\begin{figure}[ht]
\centering
\subfloat{\includegraphics[scale=.8]{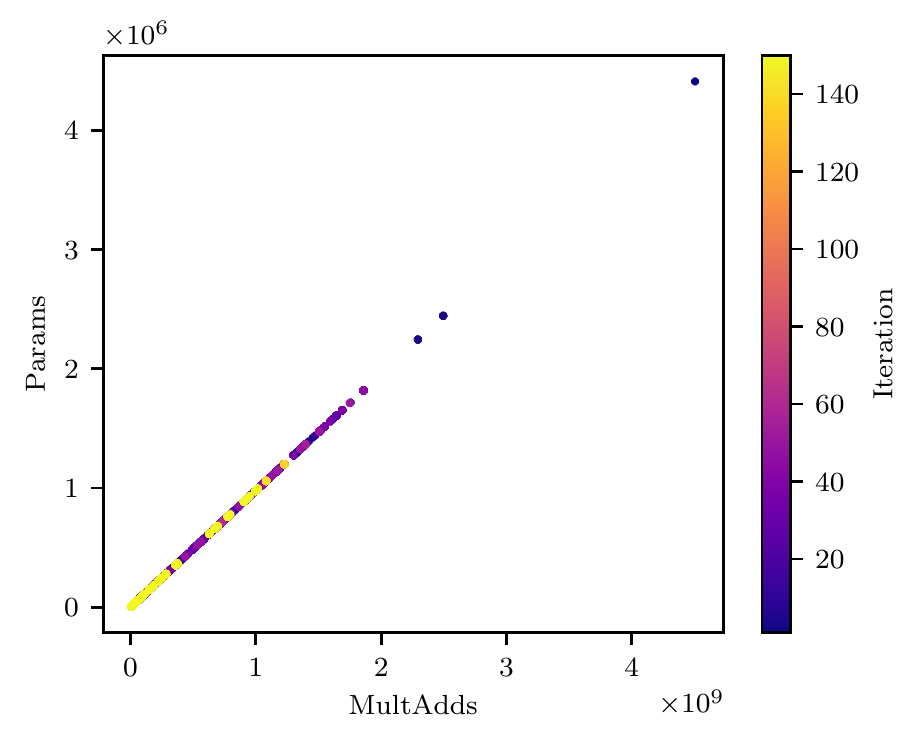}} \\
\vskip -0.1in
\subfloat{\includegraphics[scale=.8]{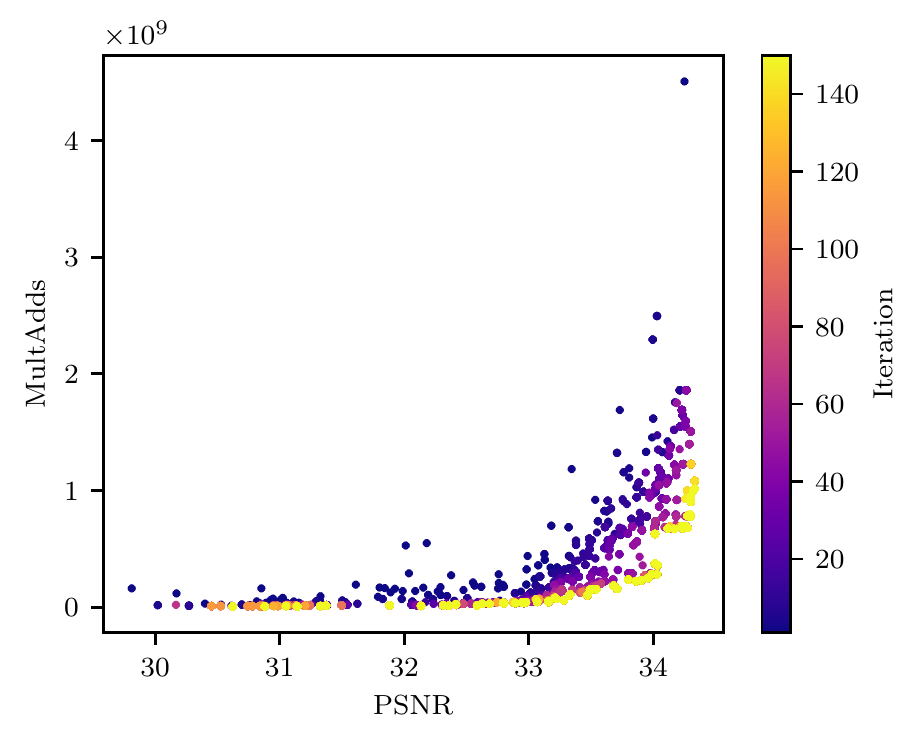}} \\
\caption{The Pareto-front of of all the models during the evolution, paired every two objectives.}
\label{fig:pareto-front}
\vskip -0.1in
\end{figure}

In our experiment, about 10k models are generated in total, where the population for each iteration is 64. The Pareto-front of all the models is shown in Fig. \ref{fig:pareto-front}. It takes less than 3 days on a Tesla-V100 with 8 GPUs to execute the pipeline once. We use DIV2K as our training set.

During an incomplete training, each model is trained with a batch size of 16 for 200 epochs. In addition, we apply Adam optimizer ($\beta_1=0.9$, $\beta_2=0.999$)  to minimize the $L_1$ loss between the generated high-resolution images and its ground truth. The learning rate is initialized as $10^{-4}$ and kept unchanged at this stage.

As for the full train, we choose 4 models with a large crowding distance in the Pareto front between mean squared error and mult-adds, which was generated at the incomplete training stage. These models are trained based on DIV2K dataset for 24000 epochs with a batch-size of 16 and it takes less than 1.5 days. Moreover, the standard deviation of weights $w$ is initialized as 0.02 and the bias 0.

\begin{table*}[t]
	\vskip 0.1in
		\caption{Comparisons with the state-of-the-art methods based on $\times$2 super-resolution task.}
	\begin{center}
				\begin{tabular}{lrrcccc}
					\toprule
					Model & Mult-Adds & Params & SET5 & SET14 & B100 & Urban100\\
					& &   &  PSNR/SSIM &  PSNR/SSIM &  PSNR/SSIM &  PSNR/SSIM \\
					\midrule
					SRCNN \cite{dong2014learning} & 52.7G & 57K & 36.66/0.9542 & 32.42/0.9063 & 31.36/0.8879 & 29.50/0.8946 \\ 
					FSRCNN \cite{dong2016accelerating} & 6.0G & 12K & 37.00/0.9558 & 32.63/0.9088 & 31.53/0.8920 & 29.88/0.9020 \\ 
					VDSR \cite{kim2016accurate} & 612.6G & 665K & 37.53/0.9587 & 33.03/0.9124 & 31.90/0.8960 & 30.76/0.9140 \\ 
					DRCN \cite{kim2016deeply}  & 17,974.3G & 1,774K & 37.63/0.9588& 33.04/0.9118& 31.85/0.8942& 30.75/0.9133 \\
					LapSRN \cite{lai2017deep} & 29.9G & 813K & 37.52/0.9590& 33.08/0.9130& 31.80/0.8950 & 30.41/0.9100 \\
					DRRN \cite{tai2017image} & 6,796.9G & 297K & 37.74/0.9591 & 33.23/0.9136 & 32.05/0.8973 & 31.23/0.9188 \\
					SelNet \cite{choi2017deep} & 225.7G & 974K & 37.89/0.9598 & 33.61/0.9160 & 32.08/0.8984 & -\\
					CARN \cite{ahn2018fast} & 222.8G & 1,592K & 37.76/0.9590 & 33.52/0.9166& 32.09/0.8978 & 31.92/0.9256\\
					CARN-M \cite{ahn2018fast} & 91.2G & 412K & 37.53/0.9583 & 33.26/0.9141 & 31.92/0.8960 & 31.23/0.9194\\
					MoreMNAS-A \cite{chu2019multi} & 238.6G & 1,039K & 37.63/0.9584 & 33.23/0.9138 & 31.95/0.8961 & 31.24/0.9187\\
					AWSRN-M \cite{wang2019lightweight} & 244.1G & 1,063K &  38.04/0.9605 & 33.66/0.9181 & 32.21/0.9000 & 32.23/0.9294 \\
					FALSR-A (ours) &234.7G & 1,021K & 37.82/0.9595 & 33.55/0.9168	& 32.12/0.8987 & 31.93/0.9256\\
					FALSR-B (ours) & 74.7G & 326k & 37.61/0.9585	& 33.29/0.9143 & 31.97/0.8967 	& 31.28/0.9191 \\
					FALSR-C (ours) & 93.7G &408k & 37.66/0.9586	& 33.26/0.9140 & 31.96/0.8965	& 31.24/0.9187 \\
					\bottomrule
				\end{tabular}
	\end{center}
		\label{tab:psnr_ssim}
\end{table*}

\begin{figure}[ht]
\vskip 0.2in
\begin{center}
\centerline{
	\begin{tikzpicture}[thick,scale=0.7, every node/.style={scale=0.7},node distance=1.5cm]
		\node (0) [darklayer] {\textcolor{white}{feature extraction}};
		\node (1) [cell, below of=0] {conv\_f64\_k3\_b4\_isskip};
		\node (2) [cell, below of=1] {conv\_f48\_k1\_b1\_isskip};
		\node (3) [cell, below of=2] {conv\_f64\_k3\_b4\_isskip};
		\node (4) [cell, below of=3] {conv\_f64\_k3\_b4\_isskip};
		\node (5) [cell, below of=4] {conv\_f64\_k3\_b4\_isskip};
		\node (6) [cell, below of=5] {conv\_f64\_k1\_b4\_noskip};
		\node (7) [cell, below of=6] {conv\_f64\_k3\_b4\_isskip};
		\node (8) [darklayer, below of=7] {\textcolor{white}{sub-pixel upsampling}};
		\draw [arrow] (0) -- (1);
		\draw [arrow] (1) -- (2);
		\draw [arrow] (2) -- (3);
		\draw [arrow] (3) -- (4);
		\draw [arrow] (4) -- (5);
		\draw [arrow] (5) -- (6);
		\draw [arrow] (6) -- (7);
		\draw [arrow] (7) -- (8);
		\newcounter{i}
		\newcommand{\drawconnections}[2]{
				\setcounter{i}{1}
				\def\s{#1}
				\foreach \j in {#2}{		    
					\stepcounter{i}	
			  		\def\a{3.6132794472845529}
			  		\def\b{0.44078070082807075}
			  		\def\c{1.8250078578729862}
			  		\pgfmathparse{\a * exp(-\b * \thei) + \c}
      			  		\xdef\looseness{\pgfmathresult}
					\pgfmathparse{int(\thei+\s)}
					\xdef\t{\pgfmathresult}
			  		\ifthenelse {\equal{\j}{0}}
						{} 
						{\draw [arrow] (#1.south) to [in=0, in looseness=\looseness, out=0, out looseness=\looseness]  (\t.north);
						} 
				}	
		};
		\drawconnections{0}{0, 0, 0, 1, 0, 1, 1};
		\drawconnections{1}{0, 0, 0, 0, 0, 1};
		\drawconnections{2}{1, 1, 0, 1, 1};
		\drawconnections{3}{0, 0, 0, 1};
		\drawconnections{4}{0, 0, 1};
		\drawconnections{5}{1, 1};
		\drawconnections{6}{1};
	\end{tikzpicture}
}
\vskip 0.1in
\caption{The model FALSR-A (the one with best visual effects) comparable to CARN. Note for instance, `conv\_f64\_k3\_b4\_isskip' represents a block of 4 convolution layers, each with a filter size of 64 and a kernel size of 3$\times$3, including a skip connection to form residual structure.}
\label{fig:FALSR-A}
\end{center}
\vskip -0.2in
\end{figure}
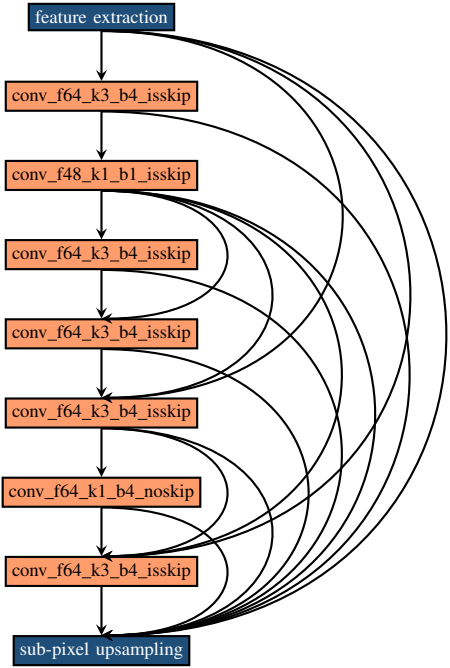

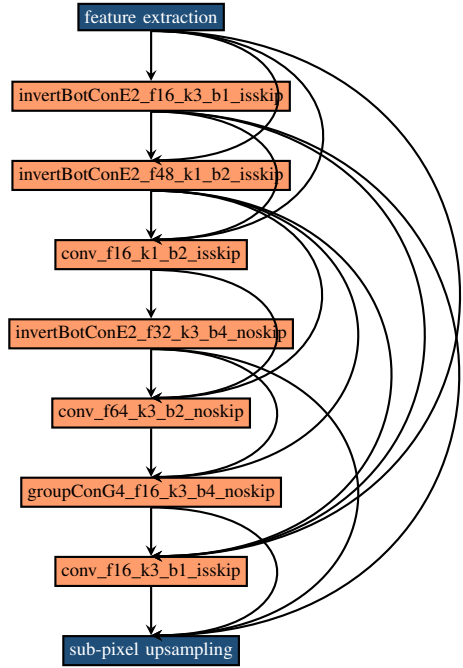
\begin{figure}[ht]
\vskip 0.2in
\begin{center}
\centerline{
	\begin{tikzpicture}[thick,scale=0.7, every node/.style={scale=0.7},node distance=1.5cm]
		\node (0) [darklayer] {\textcolor{white}{feature extraction}};
		\node (1) [cell, below of=0] {invertBotConE2\_f16\_k3\_b1\_isskip};
		\node (2) [cell, below of=1] {invertBotConE2\_f48\_k1\_b2\_isskip};
		\node (3) [cell, below of=2] {conv\_f16\_k1\_b2\_isskip};
		\node (4) [cell, below of=3] {invertBotConE2\_f32\_k3\_b4\_noskip};
		\node (5) [cell, below of=4] {conv\_f64\_k3\_b2\_noskip};
		\node (6) [cell, below of=5] {groupConG4\_f16\_k3\_b4\_noskip};
		\node (7) [cell, below of=6] {conv\_f16\_k3\_b1\_isskip};
		\node (8) [darklayer, below of=7] {\textcolor{white}{sub-pixel upsampling}};
		\draw [arrow] (0) -- (1);
		\draw [arrow] (1) -- (2);
		\draw [arrow] (2) -- (3);
		\draw [arrow] (3) -- (4);
		\draw [arrow] (4) -- (5);
		\draw [arrow] (5) -- (6);
		\draw [arrow] (6) -- (7);
		\draw [arrow] (7) -- (8);
		\newcommand{\drawconnections}[2]{
				\setcounter{i}{1}
				\def\s{#1}
				\foreach \j in {#2}{		    
					\stepcounter{i}	
			  		\def\a{3.6132794472845529}
			  		\def\b{0.44078070082807075}
			  		\def\c{1.8250078578729862}
			  		\pgfmathparse{\a * exp(-\b * \thei) + \c}
      			  		\xdef\looseness{\pgfmathresult}
					\pgfmathparse{int(\thei+\s)}
					\xdef\t{\pgfmathresult}
			  		\ifthenelse {\equal{\j}{0}}
						{} 
						{\draw [arrow] (#1.south) to [in=0, in looseness=\looseness, out=0, out looseness=\looseness]  (\t.north);
						} 
				}	
		};
		\drawconnections{0}{1, 1, 0, 0, 0, 1, 0};
		\drawconnections{1}{1, 0, 0, 0, 1, 1};
		\drawconnections{2}{0, 1, 1, 1, 0};
		\drawconnections{3}{1, 0, 0, 0};
		\drawconnections{4}{1, 0, 1};
		\drawconnections{5}{0,0};
		\drawconnections{6}{1};
		\end{tikzpicture}
}
\vskip 0.1in
\caption{The model FALSR-B comparable to CARN-M.}
\label{fig:FALSR-B}
\end{center}
\vskip -0.2in
\end{figure}

\subsection{Comparisons with State-of-the-Art Super-Resolution Methods}

\begin{figure}[ht]
\vskip 0.2in
\centering
\includegraphics[scale=.7]{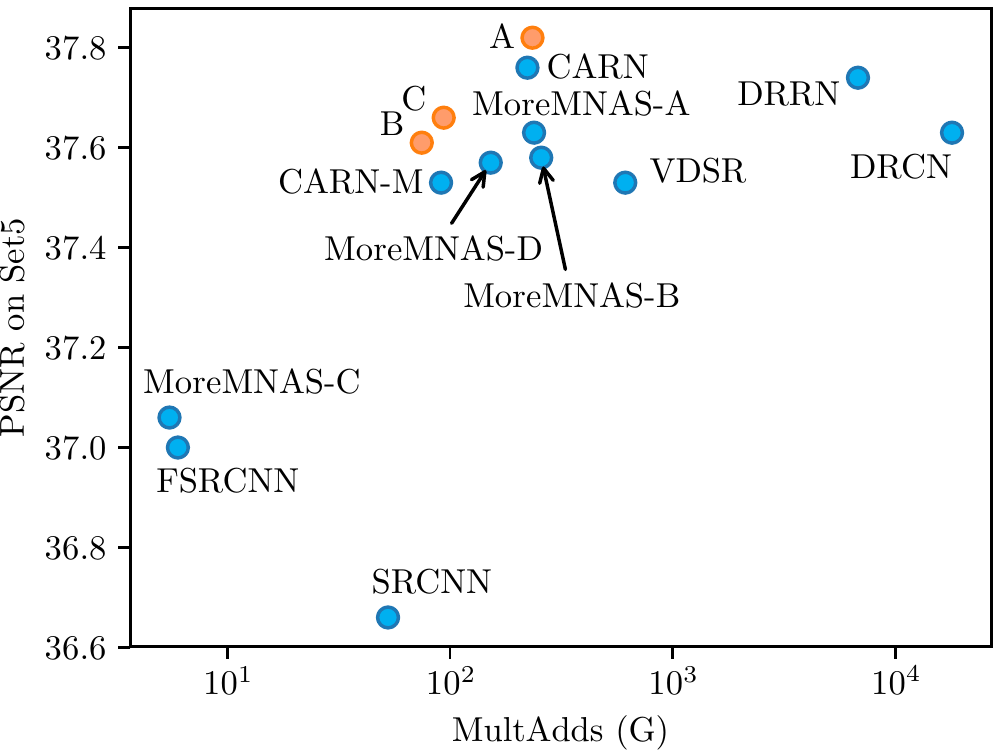}\\
\caption{FALSR A, B, C (shown in salmon) vs. others (light blue)}
\label{fig:falsr-vs-sr-set5}
\vskip -0.1in
\end{figure}

After being fully trained, our model are compared with the state-of-the-art methods on the commonly used test dataset for super-resolution (See Table \ref{tab:psnr_ssim} and Figure \ref{fig:falsr-vs-sr-set5}). To be fair, we only consider the models with comparable FLOPS. Therefore, too deep and large models such as RDN \cite{zhang2018residual}, RCAN \cite{zhang2018image} are excluded here. We choose PSNR and SSIM as metrics by convention \cite{hore2010image}. The comparisons are made on the $\times 2$ task. Note that all mult-adds are measured based on a $480\times480$ input.

At a comparable level of FLOPS, our model called FALSR-A (Figure~\ref{fig:FALSR-A}) outperforms CARN  \cite{ahn2018fast} with higher scores. In addition, it dominates DRCN \cite{kim2016deeply} and MoreMNAS-A \cite{chu2019multi} over three objectives on four datasets. Moreover, it achieves higher PSNR and SSIM with fewer FLOPS than VDSR \cite{kim2016accurate}, DRRN \cite{tai2017image} and many others.

For a more lightweight version, one model called FALSR-B (Figure~\ref{fig:FALSR-B}) dominates CARN-M, which means with fewer FLOPS and a smaller number of parameters it scores equally to or higher than CARN-M. Besides, its architecture is attractive and the complexity of connections lies in between residual and dense connections. This means a dense connection is not always the optimal way to transmit information. Useless features from lower layers could make trouble for high layers to restore super-resolution results. 

Another lightweight model called FALSR-C (not drawn because of space) also outperforms CARN-M. This model uses relatively sparse connections (8 in total). We conclude that this sparse flow works well with the selected cells. 


\begin{figure*}[ht]
\vskip 0.2in
\centering
    \subfloat[Ground Truth]{\includegraphics[scale=1.4,trim={0 30 150 120},clip]{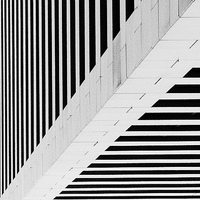}}
    \quad
     \subfloat[CARN]{\includegraphics[scale=1.4,trim={0 30 150 120},clip]{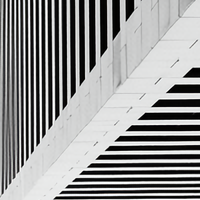}}
     \quad
     \subfloat[FALSR-A]{\includegraphics[scale=1.4,trim={0 30 150 120},clip]{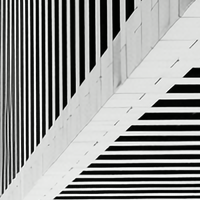}} 
     \quad
     \subfloat[CARN-M]{\includegraphics[scale=1.4,trim={0 30 150 120},clip]{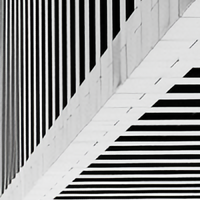}}
     \quad
     \subfloat[FALSR-B]{\includegraphics[scale=1.4,trim={0 30 150 120},clip]{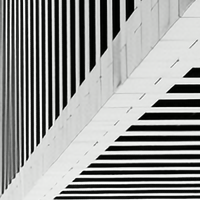}}
     \quad
     \subfloat[FALSR-C]{\includegraphics[scale=1.4,trim={0 30 150 120},clip]{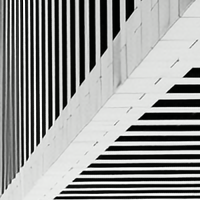}} \\
      \subfloat[Ground Truth]{\includegraphics[scale=1.4,trim={10 0 40 50},clip]{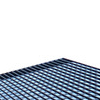}}
    \quad
     \subfloat[CARN]{\includegraphics[scale=1.4,trim={10 0 40 50},clip]{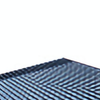}}
     \quad
     \subfloat[FALSR-A]{\includegraphics[scale=1.4,trim={10 0 40 50},clip]{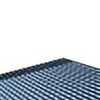}} 
     \quad
     \subfloat[CARN-M]{\includegraphics[scale=1.4,trim={10 0 40 50},clip]{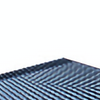}}
     \quad
     \subfloat[FALSR-B]{\includegraphics[scale=1.4,trim={10 0 40 50},clip]{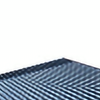}}
     \quad
     \subfloat[FALSR-C]{\includegraphics[scale=1.4,trim={10 0 40 50},clip]{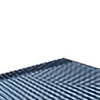}} \\
    \subfloat[Ground Truth]{\includegraphics[scale=1.4,trim={0 50 50 0},clip]{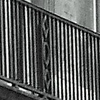}}
    \quad
     \subfloat[CARN]{\includegraphics[scale=1.4,trim={0 50 50 0},clip]{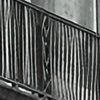}}
     \quad
     \subfloat[FALSR-A]{\includegraphics[scale=1.4,trim={0 50 50 0},clip]{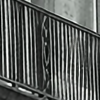}} 
     \quad
     \subfloat[CARN-M]{\includegraphics[scale=1.4,trim={0 50 50 0},clip]{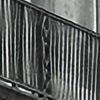}}
     \quad
     \subfloat[FALSR-B]{\includegraphics[scale=1.4,trim={0 50 50 0},clip]{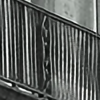}}
     \quad
     \subfloat[FALSR-C]{\includegraphics[scale=1.4,trim={0 50 50 0},clip]{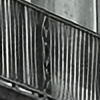}} \\   
      \subfloat[Ground Truth]{\includegraphics[scale=1.4,trim={0 20 50 30},clip]{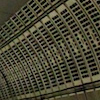}}
    \quad
     \subfloat[CARN]{\includegraphics[scale=1.4,trim={0 20 50 30},clip]{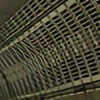}}
     \quad
     \subfloat[FALSR-A]{\includegraphics[scale=1.4,trim={0 20 50 30},clip]{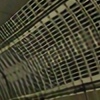}} 
     \quad
     \subfloat[CARN-M]{\includegraphics[scale=1.4,trim={0 20 50 30},clip]{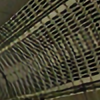}}
     \quad
     \subfloat[FALSR-B]{\includegraphics[scale=1.4,trim={0 20 50 30},clip]{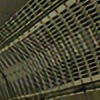}}
     \quad
     \subfloat[FALSR-C]{\includegraphics[scale=1.4,trim={0 20 50 30},clip]{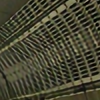}} \\
\caption{Qualitative results on images from Urban100 (image ids in rows from top to bottom: 011, 062, 066, 078).}
\label{fig:results-comparison}
\vskip -0.1in
\end{figure*}

Figure \ref{fig:results-comparison} shows the qualitative results against other methods.

\subsection{Discussions}
\subsubsection{Cell Diversity}
Our experiments show that a good cell diversity also helps to achieve better results for super-resolution, same for classification tasks \cite{hsu2018monas}. In fact, we have trained several models with repeated blocks, however, they underperform the models with diverse cells. We speculate that different types of cells can handle input features more effectively than monotonous ones.
\subsubsection{Optimal Information Flow}
Perhaps under given current technologies, dense connections are not optimal in most cases. In principle, a dense connection has the capacity to cover other non-dense configurations, however, it's usually difficult to train a model to ignore useless information.
\subsubsection{Good Assumption?}
Super-resolution is different from feature extraction domains such as classification, where more details need to be restored at pixel level. Therefore, it rarely applies downsampling operations to reduce the feature dimensions and it is more time-consuming than classification tasks like on CIFAR-10. 

Regarding the time, we use incomplete training to differentiate various models. This strategy works well under an implicit assumption: models that perform better when fully trained also behave well with a large probability under an incomplete training. Luckily, most of deep learning tasks share this good feature. For the rest, we must train models as fully as possible.

\section{Conclusions}
To sum up, we presented a novel elastic method for NAS that incorporates both micro and macro search, dealing with neural architectures in multi-granularity. The result is exciting as our generated models dominate the newest state-of-the-art SR methods. Different from human-designed and single-objective NAS models, our methods can generate different tastes of models by one run, ranging from fast and lightweight to relatively large and more accurate. Therefore, it offers a feasible way for engineers to compress existing popular human-designed models or to design various levels of architectures accordingly for constrained devices.

Our future work will focus on training a model regressor, which estimates the performance of models, to speed up the pipeline.

\bibliographystyle{IEEEtran}
\bibliography{ijcai19}

%
%
%

\end{document}